\documentclass[10pt,twocolumn,letterpaper]{article}

\usepackage{iccv}              
\usepackage{float}
\definecolor{iccvblue}{rgb}{0.21,0.49,0.74}
\usepackage[pagebackref,breaklinks,colorlinks,allcolors=iccvblue]{hyperref}
\usepackage{longtable}
\usepackage{booktabs}
\usepackage{multirow}
\usepackage{graphicx}
\usepackage{geometry}
\usepackage{array}

\def\paperID{27} 
\def\confName{ICCV}
\def\confYear{2025}

\title{Multiplicative Loss for Enhancing Semantic Segmentation \\ in Medical and Cellular Images}

\author{
Yuto Yokoi\\
Meijo University,Japan\\
253427028@ccalumni.meijo-u.ac.jp\\
\and
Kazuhiro Hotta\\
Meijo University,Japan\\
kazuhotta@meijo-u.ac.jp\\
}

\begin{document}
\maketitle
\begin{abstract}

We propose two novel loss functions, \textit{Multiplicative Loss} and \textit{Confidence-Adaptive Multiplicative Loss}, for semantic segmentation in medical and cellular images. Although Cross Entropy and Dice Loss are widely used, their additive combination is sensitive to hyperparameters and often performs suboptimally, especially with limited data.
Medical images suffer from data scarcity due to privacy, ethics, and costly annotations, requiring robust and efficient training objectives. Our \textit{Multiplicative Loss} combines Cross Entropy and Dice losses multiplicatively, dynamically modulating gradients based on prediction confidence. This reduces penalties for confident correct predictions and amplifies gradients for incorrect overconfident ones, stabilizing optimization.
Building on this, \textit{Confidence-Adaptive Multiplicative Loss} applies a confidence-driven exponential scaling inspired by Focal Loss, integrating predicted probabilities and Dice coefficients to emphasize difficult samples. This enhances learning under extreme data scarcity by strengthening gradients when confidence is low.
Experiments on cellular and medical segmentation benchmarks show our framework consistently outperforms tuned additive and existing loss functions, offering a simple, effective, and hyperparameter-free mechanism for robust segmentation under challenging data limitations.

\end{abstract}    

\section{Introduction}
\label{sec:intro}

The design of loss functions for semantic segmentation \cite{kato2022adaptivetvmfdiceloss} is a critical challenge, particularly in the medical imaging domain \cite{medicalsankou}. In medical imaging, target structures are often small and localized, while data acquisition is inherently constrained by patient privacy regulations, ethical approvals, and the substantial cost of expert annotations \cite{privercy}. As a result, the number of available training samples is fundamentally limited, making loss function design an even more decisive factor than model architecture in achieving high performance under such data-scarce conditions.

Conventional loss functions, such as Cross Entropy  \cite{crossetropysankou} and Dice Loss \cite{dicelosssankou}, have been widely adopted due to their strengths in optimizing classification accuracy and spatial overlap, respectively. However, they still suffer from several issues, including class imbalance, annotation noise, outliers, and unstable gradients. These issues are particularly pronounced in medical imaging, where extreme class imbalance and even minor annotation errors can directly degrade model performance. Therefore, there is a strong need for more stable and flexible learning objectives that can guide training robustly in these scenarios.

To address these challenges, we propose a novel loss function, \textit{Multiplicative Loss}, which multiplicatively integrates Cross Entropy Loss and Dice Loss. This design leverages the complementary properties of both losses without requiring additional hyperparameter tuning, unlike additive formulations \cite{dicetasucrosssankou}\cite{dicefocalsankou}, thus offering a highly practical and generalizable solution for real-world applications. Furthermore, we introduce the \textit{Confidence-Adaptive Multiplicative Loss} (CAML), which dynamically adjusts the loss scaling based on prediction confidence as training progresses. Specifically, CAML employs exponential scaling of the cross-entropy component according to the batch-wise average prediction confidence, enabling it to suppress overfitting as high-confidence predictions accumulate, while adaptively emphasizing hard-to-classify samples that remain uncertain.

We evaluate the proposed loss functions through experiments on two medical image datasets and one cellular image dataset with three different model architectures. As a result, the \textit{Multiplicative Loss} consistently improves both per-class IoU and mean IoU across all datasets compared to individual loss functions, and achieves comparable or superior performance to additive loss formulations. For example, on the CHASE dataset, it achieved a 2.2\% increase in mean IoU. Notably, CAML demonstrates superior performance even when the number of training samples is significantly reduced, achieving a 1.0-1.4\% improvement over conventional methods during training data reduction on the COVID-19 dataset.

The remainder of this paper is organized as follows. \cref{sec2} reviews related works. \cref{sec3} details the proposed method. \cref{sec4} presents experimental results. \cref{sec5} provides in-depth analysis. Finally, \cref{concul} concludes the paper and outlines future directions.

\section{Relate Works}
\label{sec2}
\subsection{On the Addition of Loss Functions}

A widely adopted approach in medical image segmentation is to combine Dice loss and Cross Entropy (CE) loss through a weighted summation. A representative example is the nnU-Net framework\cite{dicetasucrosssankou}, a self-adapting U-Net-based architecture that has demonstrated state-of-the-art performance on both 2D and 3D medical imaging tasks. In its training procedure, nnU-Net employs an additive loss function that integrates Dice loss and Cross Entropy loss, defined as
\begin{equation}
\mathcal{L}_{\text{total}} = \lambda \cdot \mathcal{L}_{\text{Dice}} + (1 - \lambda) \cdot \mathcal{L}_{\text{CE}},
\label{eq:additive}
\end{equation}
where $\lambda \in [0,1]$ is a hyperparameter that balances the contributions of each loss and must be carefully tuned depending on model architecture and dataset characteristics.

In this weighted additive formulation, Dice loss encourages global shape consistency and region-level overlap, while Cross Entropy loss complements pixel-wise accuracy, leading to improve the segmentation performance. However, since these two losses exhibit different gradient characteristics, their balance can become unstable as training progresses, particularly under data-constrained conditions where unstable gradients and overfitting are more likely to occur.

To address these issues and achieve higher segmentation accuracy, we propose a novel loss function, termed \textit{Confidence-Adaptive Multiplicative Loss (CAML)}. CAML dynamically adjusts the learning pressure based on prediction confidence: it responds strongly to misclassifications in the early stages of training and effectively suppresses overfitting as learning progresses. This design enables stable and accurate learning even with limited training data, making it practically applicable in the medical imaging domain.

\begin{figure*}[t]
    \centering
    \includegraphics[width=\textwidth]{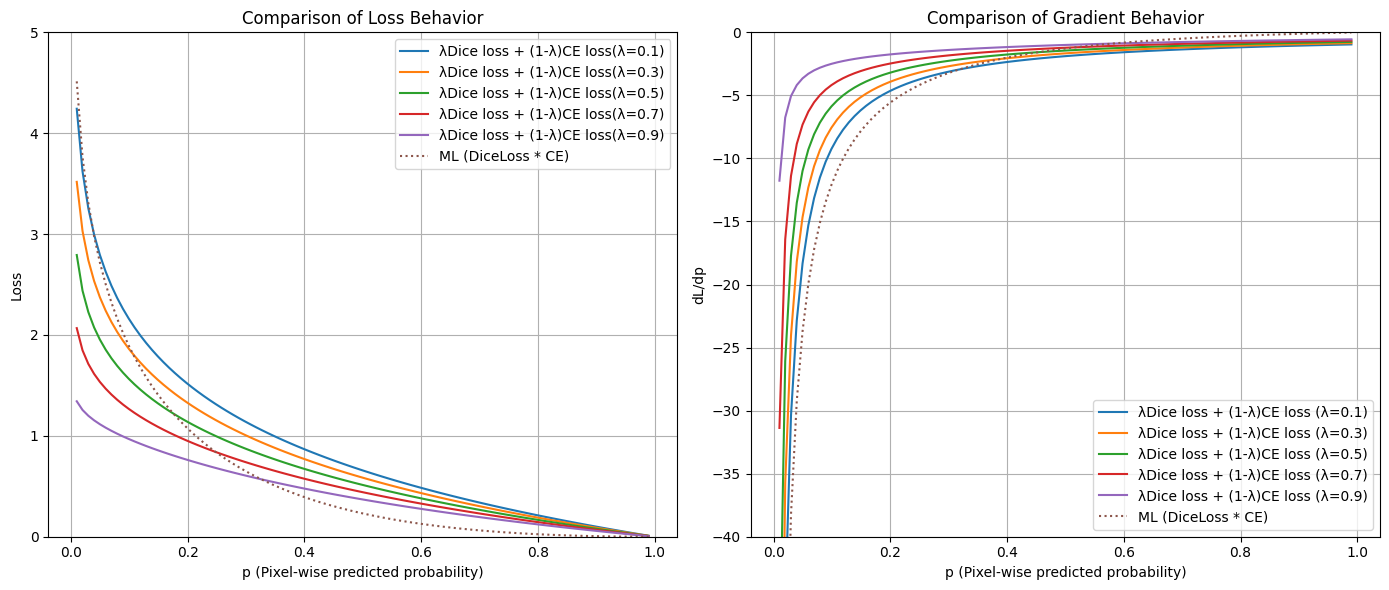}
    \caption{Visualization of loss values and gradient magnitudes with respect to predicted probability $p$. The proposed Multiplicative Loss amplifies gradients for uncertain predictions and suppresses gradients for confident predictions, thus promoting stable and efficient optimization.
    }
    \label{multihikaku}
\end{figure*}


\section{Proposed Method}
\label{sec3}
\subsection{Multiplicative Loss}

In this section, we present the formulation and rationale of the proposed Multiplicative Loss (ML), designed to enhance semantic segmentation performance by simultaneously leveraging pixel-wise and region-level supervision. Semantic segmentation requires the model to jointly capture fine-grained local details and global spatial structures. Although conventional loss functions such as Cross Entropy Loss and Dice Loss have been widely utilized, each possesses distinct advantages and limitations.

The CE Loss is formally defined as 
\begin{equation}
    \mathcal{L}_{\mathrm{CE}} = -\frac{1}{N} \sum_{i=1}^{N} \sum_{c=1}^{C} y_{i,c} \log(p_{i,c}),
    \label{eq:ce_loss}
\end{equation}
where $N$ is batch size, $C$ is the number of classes, $y_{i,c}$ denotes one-hot ground truth for sample $i$ and class $c$, and $p_{i,c}$ represents the predicted class probability.

Dice Loss which emphasizes spatial overlap between predicted and ground truth masks is defined as 
\begin{align}
    \mathrm{Dice}_c &= \frac{2\sum_{i=1}^{N} y_{i,c} p_{i,c}}{\sum_{i=1}^{N} y_{i,c} + \sum_{i=1}^{N} p_{i,c}},
    \label{eq:dice_loss_1} \\
    \mathcal{L}_{\mathrm{Dice}} &= 1 - \frac{1}{C} \sum_{c=1}^{C} \mathrm{Dice}_c,
    \label{eq:dice_loss_2}
\end{align}
where $y_{i,c}$ and $p_{i,c}$ denote the ground truth label (one-hot encoded) and the predicted probability for class $c$ at sample $i$, respectively.

Although both losses capture complementary aspects of segmentation quality, simple additive combinations often require delicate hyperparameter tuning and may fail to fully exploit their mutual strengths. To address this, we propose a multiplicative integration defined as 
\begin{equation}
    \mathcal{L}_{\mathrm{ML}} = \mathcal{L}_{\mathrm{Dice}} \times \mathcal{L}_{\mathrm{CE}}.
    \label{eq:ml_loss}
\end{equation}
The gradient of the proposed ML loss can be derived using the product rule as shown in  \cref{eq:ml_gradient_1},\cref{eq:ml_gradient_2}
{\footnotesize
\begin{align}
    \frac{\partial \mathcal{L}_{\mathrm{ML}}}{\partial p_{i,c}} 
    &= \mathcal{L}_{\mathrm{Dice}} \cdot \frac{\partial \mathcal{L}_{\mathrm{CE}}}{\partial p_{i,c}} + 
         \mathcal{L}_{\mathrm{CE}} \cdot \frac{\partial \mathcal{L}_{\mathrm{Dice}}}{\partial p_{i,c}}
    \label{eq:ml_gradient_1} \\
    &= \mathcal{L}_{\mathrm{Dice}} \cdot (p_{i,c} - y_{i,c}) \nonumber \\
    &\quad - \mathcal{L}_{\mathrm{CE}} \cdot \frac{1}{C} \cdot \frac{2 \left( y_{i,c} \sum_{i=1}^{N} p_{i,c} - \sum_{i=1}^{N} p_{i,c} y_{i,c} \right)}{\left( \sum_{i=1}^{N} p_{i,c} + \sum_{i=1}^{N} y_{i,c} \right)^2}
    \label{eq:ml_gradient_2}
\end{align}
}
For comparison, the gradient of the conventional weighted additive loss, incorporating a trade-off parameter $\lambda$, is given by \cref{eq:additive_gradient_1},\cref{eq:additive_gradient_2}
{\footnotesize
\begin{align}
    \frac{\partial \mathcal{L}_{\mathrm{total}}}{\partial p_{i,c}} 
    &= \lambda \cdot \frac{\partial \mathcal{L}_{\mathrm{CE}}}{\partial p_{i,c}} 
    + (1 - \lambda) \cdot \frac{\partial \mathcal{L}_{\mathrm{Dice}}}{\partial p_{i,c}}
    \label{eq:additive_gradient_1} \\
    &= \lambda \cdot (p_{i,c} - y_{i,c}) \nonumber \\
    &\quad - (1 - \lambda) \cdot \frac{1}{C} \cdot 
    \frac{2 \left( y_{i,c} \sum_{i=1}^{N} p_{i,c} - \sum_{i=1}^{N} p_{i,c} y_{i,c} \right)}{\left( \sum_{i=1}^{N} p_{i,c} + \sum_{i=1}^{N} y_{i,c} \right)^2}
    \label{eq:additive_gradient_2}
\end{align}
}

Unlike the additive scheme that requires manual tuning of $\lambda$ to balance the two loss terms, our Multiplicative Loss adaptively modulates the contribution of each component based on their instantaneous values during training. This leads to two key advantages: (i) stronger gradients for misclassified or uncertain predictions (low confidence), facilitating accelerated learning on hard examples, and (ii) attenuated gradients for highly confident predictions, which mitigates overfitting as training progresses.

These behaviors are visually illustrated in \cref{multihikaku}. Graphs show how the proposed Multiplicative Loss amplifies gradients when predictions are uncertain and suppresses them when predictions are high confident, thereby promoting stable and efficient optimization.

In summary, the proposed Multiplicative Loss offers an elegant, hyperparameter-free formulation that enables robust convergence without compromising segmentation accuracy, especially in data-limited scenarios common in medical imaging applications.

\begin{figure*}[h]
    \centering
    \includegraphics[width=\textwidth]{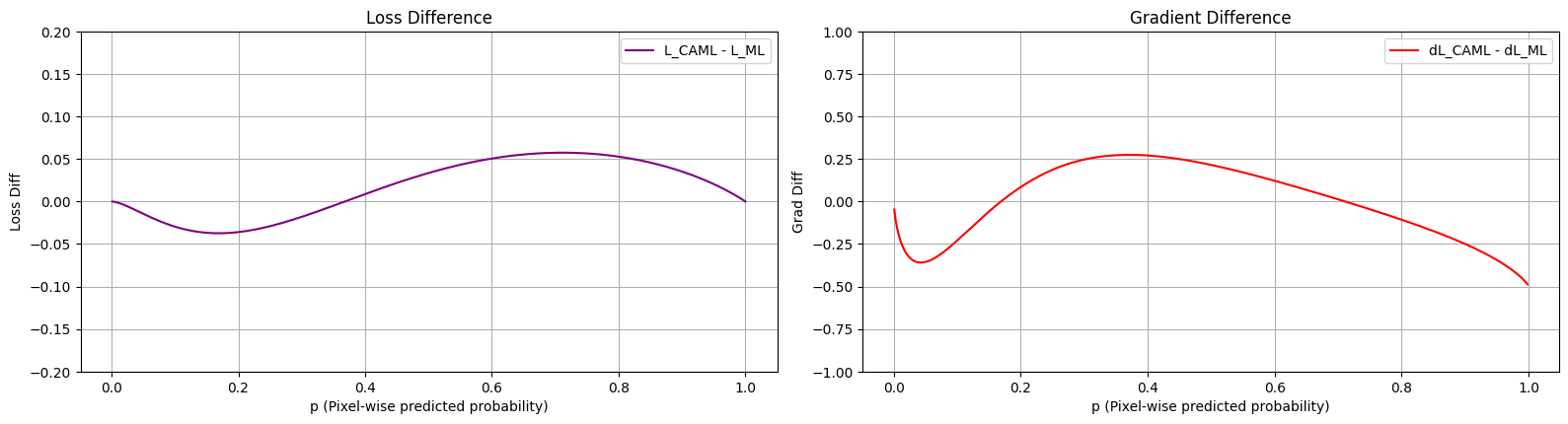}
    \caption{Comparison of gradient magnitudes between CAML and ML with respect to the predicted probability $p$. CAML adaptively adjusts gradient strength based on instance-level confidence, emphasizing hard samples while suppressing excessive updates for high-confidence predictions.}
    \label{camlhikaku}
\end{figure*}

\subsection{Confidence-Adaptive Multiplicative Loss}
\label{3.2}

In this section, we extend the proposed ML introduced in \cref{eq:ml_loss} by incorporating an adaptive weighting mechanism that dynamically adjusts the loss contribution based on per-pixel prediction confidence through training. This extension called CAML is defined as
{\footnotesize
\begin{equation}
    \mathcal{L}_{\mathrm{CAML}} = \mathcal{L}_{\mathrm{Dice}} \cdot \left( \mathcal{L}_{\mathrm{CE}} \right)^{\alpha}, \quad \text{where} \quad \alpha = (1 - p)^D,
    \label{eq:CAML_compact}
\end{equation}
}
where $\mathcal{L}_{\mathrm{Dice}}$ and $\mathcal{L}_{\mathrm{CE}}$ denote the Dice loss and Cross Entropy loss (see \cref{eq:dice_loss_2} and \cref{eq:ce_loss}), respectively. $p$ represents the predicted probability for the true class at each pixel, and $D$ is the Dice coefficient in \cref{eq:dice_loss_1} indicating segmentation consistency.

The gradient of $\mathcal{L}_{\mathrm{CAML}}$ with respect to $p$ is derived via the chain rule as
{\footnotesize
\begin{equation}
\begin{aligned}
    \frac{\partial \mathcal{L}_{\mathrm{CAML}}}{\partial p} 
    =\,& \left(\mathcal{L}_{\mathrm{CE}}\right)^{\alpha} \Biggl[ 
    \frac{\partial \mathcal{L}_{\mathrm{Dice}}}{\partial p} \\
    & + \mathcal{L}_{\mathrm{Dice}} \cdot \left(
    \frac{\partial \alpha}{\partial p} \log \mathcal{L}_{\mathrm{CE}} 
    + \alpha \cdot \frac{1}{\mathcal{L}_{\mathrm{CE}}} \frac{\partial \mathcal{L}_{\mathrm{CE}}}{\partial p}
    \right) 
    \Biggr],
\end{aligned}
\label{eq:CAML_grad_compact}
\end{equation}
}
where
{\footnotesize
\begin{equation}
    \frac{\partial \alpha}{\partial p} = -D (1 - p)^{D-1}.
    \label{eq:alpha_deriv}
\end{equation}
}
by incorporating the Dice coefficient into the exponent $\alpha$, CAML dynamically modulates pixel-wise supervision based on current segmentation performance. As training progresses and both $p$ and $D$ increase, the weight on $\mathcal{L}_{\mathrm{CE}}$ decreases exponentially, effectively suppressing overfitting on easy samples while emphasizing harder cases.

Moreover, as illustrated in \cref{camlhikaku}, 
CAML adaptively scales gradient magnitudes with respect to both pixel-wise confidence and overall batch performance. For pixels with high confidence $p$, CAML’s gradients closely approximate those of ML, minimizing unnecessary updates and facilitating stable convergence.

Importantly, the adaptive mechanism of CAML is especially effective when training data is limited. In such cases, conventional additive or simple multiplicative losses tend to overfit easy samples due to the scarcity of training data.
CAML mitigates this by prioritizing harder samples, thereby balancing the bias-variance trade-off and achieving better generalization even with small datasets.

CAML mitigates this by prioritizing harder samples, thereby balancing the bias-variance trade-off and achieving better generalization even with small datasets. This effect stems from the adaptive weighting mechanism governed by the per-pixel confidence $p$ and the global Dice coefficient $D$, which together determine the exponent $\alpha = (1 - p)^D$. As training progresses and segmentation performance improves, $\alpha$ decreases for confident predictions, reducing their influence on the overall loss. This dynamic attenuation effectively suppresses gradient updates from easy pixels that are already well-learned, allowing the model to concentrate on more uncertain or challenging regions. Such targeted learning prevents overfitting to redundant information and noise, which are common issues in limited data regimes, and instead promotes better feature abstraction by focusing model capacity on informative errors. Additionally, the small but non-zero gradients in high-confidence regions serve as a form of implicit regularization, contributing to optimization stability and further enhancing generalization.


In summary, CAML enables dynamic loss adaptation aligned with training dynamics, promoting robust optimization even under limited data scenarios. Experiments on multiple medical image segmentation benchmarks demonstrate that CAML consistently outperforms conventional additive and multiplicative loss formulations, achieving superior accuracy, enhanced stability, and improved overfitting resistance. These results establish CAML as an effective loss function for small-scale medical image learning tasks.

\section{Experiments}
\label{sec4}

\subsection{Datasets and Experimental Setup}
\label{41}

In this study, we conducted comprehensive evaluations on two medical imaging datasets and one cellular imaging dataset.
The first dataset is the \textit{Drosophila} dataset\cite{haesankou}, which consists of cellular microscopy images of \textit{Drosophila} (fruit fly). This dataset includes five semantic classes: membrane, mitochondria, synapse, glia/extracellular, and intracellular. It contains a total of 20 images with a resolution of 1024 $\times$ 1024 pixels. Due to GPU memory limitations, each image was randomly cropped into 256 $\times$ 256 patches. We prepared 48 training patches (multiple random crops per image), 16 validation patches (each image divided into 16 non-overlapping patches), and 16 test patches (divided in the same manner). The performance was evaluated using five-fold cross-validation.

The second dataset is the COVID-19 lung CT dataset\cite{COVID19}, consisting of CT scans from 40 patients. The dataset includes four semantic classes: background, lungs-other, ground glass, and consolidations. All images were resized to 256 $\times$ 256 pixels. The dataset was split into 70 training images, 10 validation images, and 20 test images. Performance evaluation was conducted with five-fold cross-validation.

The third dataset is the CHASE dataset\cite{ganteisankou}, which targets retinal vessel segmentation from fundus images. It contains 28 images with a resolution of 999 $\times$ 960 pixels. Due to GPU memory constraints, all images were resized to 256 $\times$ 256 pixels. The dataset was divided into 22 for training, 6 for validation, and 6 for test, and evaluated by five-fold cross-validation.

To further assess the robustness under limited data conditions, we conducted additional experiments by reducing the number of training samples with fixed random seeds. Specifically, for both the \textit{Drosophila} and COVID-19 datasets, training was repeated with 1/4 and 1/8 of the original training data. For the CHASE dataset, due to its smaller size, experiments were conducted using 1/2 and 1/4 of the original training images. These settings enabled us to evaluate the performance of our method under small data regimes.

Three model architectures were employed in all experiments: U-Net \cite{Unet}, TransUNet \cite{transUnet}, and ConvFormer+U-Net \cite{convformer}, where ConvFormer was used as the encoder. U-Net and TransUNet were chosen for their strong performance in medical image segmentation tasks. ConvFormer+U-Net was selected to evaluate the effectiveness of convolution-based token mixing in place of attention mechanisms, as well as its demonstrated high accuracy in image classification tasks. To purely compare the effects of loss functions, all models were trained from scratch without any pre-training.

Through all experiments, the batch size was fixed at 4, and the learning rate was set to 1e-3. The Adam\cite{adam} optimizer was used for training. Data augmentation included random cropping and random horizontal flipping. All training was conducted on a single NVIDIA A6000 GPU.

For quantitative evaluation, Intersection over Union (IoU)\cite{IoU} was adopted as the primary metric. Class-wise IoU and mean IoU (mIoU) across all classes were computed. Each training setting was repeated three times with different random initializations, and the average mIoU was reported as the final evaluation score.

\subsection{Quantitative Evaluation Results}
\label{42}

We report quantitative results on three benchmark datasets: COVID-19, CHASE, and Drosophila in Tables~\cref{tab:covid-19segewa}, \cref{tab:CHASEseg}, and \cref{tab:Drosophilaseg}, respectively.

\vspace{1mm}
\noindent\textbf{COVID-19 Dataset.} On the COVID-19 dataset, the proposed ML achieved comparable or superior mIoU scores across all models compared to additive loss baselines. For instance, in the U-Net model, the IoU for the \textit{pleural effusions} class improves from 0.15\% to 1.87\%. Although ML slightly underperforms in the TransUNet model by approximately 0.2\%, this can be considered within the margin of error, suggesting no hyperparameter tuning is necessary.

CAML further improved performance, especially under limited data scenarios. When  training set is reduced to 1/8 of its original size, CAML improved mIoU by 0.86\% in U-Net, with a significant 2.53\% increase for \textit{ground glass} class. In the ConvFormer+U-Net model, CAML improves the mIoU by 1.0\% and the IoU for \textit{ground glass} by 0.8\%. However, CAML leads to a 0.4\% drop in mIoU in TransUNet when using only 1/4 of the training samples.

\vspace{1mm}
\noindent\textbf{CHASE Dataset.} For the CHASE dataset, ML improved mIoU over additive losses for all models except TransUNet. The U-Net model shows an improvement of 1.15\%–1.77\%, while ConvFormer+U-Net gains 2.20\% in mIoU, and up to 3.93\% in IoU for the \textit{retinal vessel} class.

CAML consistently improved mIoU under reduced training samples (1/4 of full set) in all models except for TransUNet. In U-Net, CAML increases mIoU by 1.20\% and improves the IoU for \textit{retinal vessel} by 1.8\%. For ConvFormer+U-Net, mIoU improves by 0.54\% and the IoU for \textit{retinal vessel} by 0.4\%.

\vspace{1mm}
\noindent\textbf{Drosophila Dataset.} In the Drosophila dataset, ML improved mIoU in all models except TransUNet. For U-Net, the improvement ranges from 0.14\% to 0.47\%. Per-class IoUs show negligible differences (up to 0.33\%), supporting that ML performs robustly without hyperparameter tuning. ConvFormer+U-Net also shows a 0.34\% increase in mIoU, with per-class IoUs matching or exceeding those of additive losses.

Under the reduced data settings, CAML outperformed additive losses in all models except for TransUNet. When we use only 1/8 of the training data, CAML improves mIoU in U-Net by 1.47\%, with a 2.0\% increase for both the \textit{mitochondria} and \textit{synapse} classes.

\vspace{1mm}
\noindent\textbf{Discussion.} These results validate the robustness and effectiveness of the proposed ML and CAML. As described in \cref{eq:ml_gradient_1}, ML computes a weighted sum of gradients from distinct loss functions, which eliminates the need for dataset-specific hyperparameter tuning while maintaining high accuracy.
Frthermore, as shown in \cref{eq:ml_loss} and \cref{eq:ml_gradient_1}, the multiplicative form of ML suppresses large parameter updates unless both loss components provide meaningful gradients. This selective update mechanism promotes focused learning on semantically important regions such as lesions or organs in medical images even if these regions are infrequent or underrepresented, thereby enabling high accuracy in such local areas.

CAML, defined in \cref{eq:CAML_compact}, further enhances this behavior. According to \cref{eq:alpha_deriv}, CAML attenuates gradients from high-confidence predictions while amplifying those from low-confidence pixels. This dynamic adjustment mitigates overfitting under limited supervision and ensures efficient learning, even when the predicted probabilities become extreme (close to 0 or 1) due to small training data sizes (see \cref{camlhikaku}).


\begin{table*}[t]
\centering
\tiny
\caption{Segmentation performance (IoU, \%) for different class and loss functions on the COVID-19 dataset.}
\label{tab:covid-19segewa}
\begin{tabular}{lccccccc}
\toprule
Networks & Loss & mIoU  & background & consolidations  & ground glass  & pleural effusions \\
\midrule

U\text{-}Net (Full data) & CE Loss & 42.88$\pm$0.61 & 96.26$\pm$0.23 & 31.53$\pm$1.15 & 42.40$\pm$1.50 & 1.40$\pm$0.61 \\
                         & Dice Loss & 44.86$\pm$1.15 & 96.20$\pm$0.00 & 32.98$\pm$2.57 & 43.63$\pm$1.50 & 7.00$\pm$1.03 \\
                         & $\lambda$Dice + (1\!-\!$\lambda$)CE ($\lambda$=0.3) & 44.53$\pm$0.94 & 96.26$\pm$0.11 & 32.06$\pm$1.51 & 44.40$\pm$1.40 & 5.53$\pm$0.90 \\
                         & $\lambda$Dice + (1\!-\!$\lambda$)CE ($\lambda$=0.5) & 44.60$\pm$0.91 & 96.40$\pm$0.20 & 33.60$\pm$1.03 & 41.66$\pm$0.98 & 6.85$\pm$1.86 \\
                         & $\lambda$Dice + (1\!-\!$\lambda$)CE ($\lambda$=0.7) & 44.40$\pm$0.20 & 96.06$\pm$0.11 & 33.80$\pm$1.40 & 42.73$\pm$1.62 & 5.13$\pm$0.83 \\
                         & ML (ours) & 45.26$\pm$0.11 & 96.53$\pm$0.11 & 33.06$\pm$0.94 & 44.46$\pm$0.70 & \textcolor{red}{7.00$\pm$0.91} \\
                         & CAML (ours) & \textcolor{red}{45.53$\pm$0.57} & \textcolor{red}{96.53$\pm$0.11} & \textcolor{red}{34.93$\pm$0.57} & \textcolor{red}{44.66$\pm$1.50} & 6.53$\pm$1.52 \\
\midrule
U\text{-}Net (1/4 Training data) & CE Loss & 39.80$\pm$0.80 & 95.46$\pm$0.23 & 25.80$\pm$2.69 & 37.60$\pm$0.60 & 0.13$\pm$0.80 \\
                                 & Dice Loss & 40.66$\pm$0.11 & 94.86$\pm$0.11 & 27.46$\pm$0.11 & 36.40$\pm$0.20 & \textcolor{red}{3.80$\pm$0.34} \\
                                 & $\lambda$Dice + (1\!-\!$\lambda$)CE ($\lambda$=0.5) & 40.26$\pm$0.57 & 95.26$\pm$0.11 & 24.93$\pm$1.72 & 37.53$\pm$1.10 & 3.06$\pm$0.57 \\
                                 & ML (ours) & 40.60$\pm$0.40 & 95.33$\pm$0.11 & 26.66$\pm$0.30 & 39.40$\pm$1.58 & 0.73$\pm$0.80 \\
                                 & CAML (ours) & \textcolor{red}{41.33$\pm$0.57} & \textcolor{red}{95.66$\pm$0.11} & \textcolor{red}{27.66$\pm$1.41} & \textcolor{red}{40.46$\pm$0.98} & 1.60$\pm$1.70 \\
\midrule
U\text{-}Net (1/8 Training data) & CE Loss & 36.46$\pm$0.30 & 94.33$\pm$0.11 & 16.93$\pm$0.61 & 36.40$\pm$0.60 & 0.00$\pm$0.30 \\
                                 & Dice Loss & 36.86$\pm$0.11 & 93.66$\pm$0.23 & 16.73$\pm$0.41 & 34.60$\pm$0.20 & \textcolor{red}{2.26$\pm$0.64} \\
                                 & $\lambda$Dice + (1\!-\!$\lambda$)CE ($\lambda$=0.5) & 37.20$\pm$0.34 & 94.26$\pm$0.41 & 19.06$\pm$1.10 & 34.33$\pm$1.00 & 1.06$\pm$1.06 \\
                                 & ML (ours) & 37.13$\pm$0.61 & 94.33$\pm$0.30 & 18.66$\pm$2.02 & 35.33$\pm$0.61 & 0.06$\pm$0.11 \\
                                 & CAML (ours) & \textcolor{red}{38.06$\pm$0.41} & \textcolor{red}{94.60$\pm$0.20} & \textcolor{red}{20.60$\pm$1.70} & \textcolor{red}{36.86$\pm$0.11} & 0.06$\pm$0.11 \\
\midrule
TransUNet (Full Training data) & CE Loss & 40.60$\pm$0.87 & 95.93$\pm$0.11 & 26.20$\pm$1.21 & 39.86$\pm$3.17 & 0.00$\pm$0.00 \\
                               & Dice Loss & 42.26$\pm$0.46 & \textcolor{red}{96.13$\pm$0.30} & \textcolor{red}{30.06$\pm$0.94} & 40.80$\pm$0.91 & 2.46$\pm$0.30 \\
                               & $\lambda$Dice + (1\!-\!$\lambda$)CE ($\lambda$=0.5) & \textcolor{red}{42.46$\pm$0.57} & 95.80$\pm$0.20 & 27.17$\pm$3.87 & 41.20$\pm$2.25 & \textcolor{red}{3.00$\pm$2.10} \\
                               & ML (ours) & 42.26$\pm$0.23 & 95.93$\pm$0.11 & 29.26$\pm$0.11 & 42.00$\pm$1.58 & 1.26$\pm$1.85 \\
                               & CAML (ours) & 42.06$\pm$0.23 & 96.06$\pm$0.11 & 29.33$\pm$0.50 & \textcolor{red}{42.00$\pm$1.38} & 1.00$\pm$1.11 \\
\midrule
TransUNet (1/4 Training data) & CE Loss & 38.33$\pm$0.30 & 95.00$\pm$0.20 & 22.53$\pm$2.08 & 35.53$\pm$0.41 & 0.00$\pm$0.00 \\
                              & Dice Loss & 39.06$\pm$0.41 & 94.60$\pm$0.20 & 23.73$\pm$1.33 & 35.33$\pm$1.20 & \textcolor{red}{2.40$\pm$0.91} \\
                              & $\lambda$Dice + (1\!-\!$\lambda$)CE ($\lambda$=0.5) & 38.26$\pm$0.41 & 94.73$\pm$0.23 & 23.13$\pm$0.64 & 36.00$\pm$0.52 & 0.20$\pm$0.41 \\
                              & ML (ours) & 39.06$\pm$0.50 & 94.93$\pm$0.41 & 23.13$\pm$0.83 & \textcolor{red}{37.33$\pm$0.64} & 0.86$\pm$1.33 \\
                              & CAML (ours) & \textcolor{red}{39.06$\pm$0.23} & \textcolor{red}{95.00$\pm$0.20} & \textcolor{red}{26.93$\pm$0.80} & 36.40$\pm$0.34 & 0.60$\pm$0.72 \\
\midrule
TransUNet (1/8 Training data) & CE Loss & 36.00$\pm$0.34 & 94.40$\pm$0.20 & 14.93$\pm$1.41 & 33.53$\pm$0.53 & 0.00$\pm$0.00 \\
                              & Dice Loss & 37.33$\pm$0.23 & 94.33$\pm$0.34 & 18.00$\pm$0.87 & 34.80$\pm$0.52 & \textcolor{red}{2.06$\pm$0.11} \\
                              & $\lambda$Dice + (1\!-\!$\lambda$)CE ($\lambda$=0.5) & \textcolor{red}{37.46$\pm$0.23} & 94.66$\pm$0.23 & \textcolor{red}{19.20$\pm$0.34} & \textcolor{red}{36.20$\pm$0.34} & 0.40$\pm$0.69 \\
                              & ML (ours) & 36.93$\pm$0.41 & \textcolor{red}{94.73$\pm$0.30} & 17.40$\pm$0.87 & 35.86$\pm$0.80 & 0.00$\pm$0.00 \\
                              & CAML (ours) & 37.00$\pm$0.00 & 94.53$\pm$0.50 & 17.40$\pm$0.69 & 35.73$\pm$1.10 & 0.33$\pm$0.57 \\
\midrule

ConvFormer+U\text{-}Net (Full data) & CE Loss & 44.06$\pm$0.50 & 96.60$\pm$0.23 & 32.80$\pm$1.58 & 45.33$\pm$0.80 & 1.66$\pm$0.30 \\
                            & Dice Loss & 44.80$\pm$0.87 & 96.20$\pm$0.34 & 34.13$\pm$1.10 & 44.06$\pm$0.83 & 5.00$\pm$0.91 \\
                            & $\lambda$Dice + (1\!-\!$\lambda$)CE ($\lambda$=0.5) & 45.13$\pm$0.70 & 96.53$\pm$0.11 & 34.06$\pm$2.23 & 43.93$\pm$0.80 & \textcolor{red}{6.26$\pm$1.17} \\
                            & ML (ours) & \textcolor{red}{45.46$\pm$0.30} & 96.53$\pm$0.11 & \textcolor{red}{34.53$\pm$1.67} & 45.13$\pm$0.30 & 6.14$\pm$0.30 \\
                            & CAML (ours) & 44.80$\pm$0.00 & \textcolor{red}{96.70$\pm$0.11} & 34.50$\pm$0.80 & \textcolor{red}{46.20$\pm$1.89} & 1.70$\pm$1.70 \\
\midrule
ConvFormer+U\text{-}Net (1/4 Training data) & CE Loss & 40.46$\pm$0.50 & \textcolor{red}{95.80$\pm$0.20} & 26.13$\pm$1.47 & \textcolor{red}{39.53$\pm$1.41} & 0.00$\pm$0.00 \\
                                    & Dice Loss & \textcolor{red}{41.06$\pm$0.83} & 95.06$\pm$0.30 & 27.80$\pm$1.44 & 37.93$\pm$1.20 & \textcolor{red}{3.46$\pm$1.94} \\
                                    & $\lambda$Dice + (1\!-\!$\lambda$)CE ($\lambda$=0.5) & 40.66$\pm$0.50 & 95.33$\pm$0.30 & \textcolor{red}{29.93$\pm$3.71} & 38.46$\pm$1.26 & 0.86$\pm$1.02 \\
                                    & ML (ours) & 40.00$\pm$0.34 & 95.46$\pm$0.11 & 25.06$\pm$3.52 & 38.40$\pm$1.38 & 0.80$\pm$1.05 \\
                                    & CAML (ours) & 39.80$\pm$0.87 & 95.33$\pm$0.11 & 25.13$\pm$1.50 & 36.86$\pm$2.80 & 1.40$\pm$0.52 \\
\midrule
ConvFormer+U\text{-}Net (1/8 Training data) & CE Loss & 36.46$\pm$0.30 & 94.00$\pm$0.40 & 18.20$\pm$2.25 & 32.86$\pm$1.27 & 0.00$\pm$0.00 \\
                                    & Dice Loss & 38.00$\pm$0.87 & 94.00$\pm$0.34 & 22.13$\pm$2.85 & 33.66$\pm$1.47 & \textcolor{red}{2.13$\pm$0.11} \\
                                    & $\lambda$Dice + (1\!-\!$\lambda$)CE ($\lambda$=0.5) & 37.53$\pm$0.30 & 93.93$\pm$0.30 & 19.06$\pm$1.81 & 35.40$\pm$1.11 & 1.26$\pm$0.80 \\
                                    & ML (ours) & 37.13$\pm$0.23 & 94.26$\pm$0.23 & 19.73$\pm$1.15 & 34.80$\pm$1.50 & 0.00$\pm$0.00 \\
                                    & CAML (ours) & \textcolor{red}{38.53$\pm$1.13} & \textcolor{red}{94.60$\pm$0.20} & \textcolor{red}{22.20$\pm$1.92} & \textcolor{red}{36.20$\pm$2.25} & 0.40$\pm$0.69 \\
\bottomrule
\end{tabular}
\end{table*}

\begin{table*}[t]
\centering
\tiny
\caption{Segmentation performance (IoU, \%) for different class and loss functions on the CHASE dataset.}
\label{tab:CHASEseg}
\begin{tabular}{lccccc}
\toprule
Networks & Loss & mIoU  & background & retinal vessel \\
\midrule

U\text{-}Net (Full data) & CE Loss & 60.13$\pm$1.51 & 93.53$\pm$0.30 & 26.66$\pm$2.87 \\
                  & Dice Loss & 62.90$\pm$0.90 & 92.53$\pm$0.61 & \textcolor{red}{33.06$\pm$1.50} \\
                  & $\lambda$Dice loss + (1\!-\!$\lambda$)CE loss ($\lambda$=0.3) & 61.66$\pm$1.55 & 93.53$\pm$0.30 & 29.93$\pm$2.91 \\
                  & $\lambda$Dice loss + (1\!-\!$\lambda$)CE loss ($\lambda$=0.5) & 61.33$\pm$1.22 & 92.93$\pm$0.30 & 29.73$\pm$2.21 \\
                  & $\lambda$Dice loss + (1\!-\!$\lambda$)CE loss ($\lambda$=0.7) & 61.86$\pm$1.85 & 92.86$\pm$0.70 & 30.66$\pm$3.28 \\
                  & ML (ours) & 62.80$\pm$0.87 & 93.20$\pm$0.00 & 32.26$\pm$1.33 \\
                  & CAML (ours) & \textcolor{red}{63.00$\pm$1.40} & \textcolor{red}{93.53$\pm$0.30} & 32.86$\pm$2.77 \\
\midrule
U\text{-}Net (1/2 Training data) & CE Loss & 59.06$\pm$0.83 & \textcolor{red}{93.66$\pm$0.23} & 24.60$\pm$1.70 \\
                         & Dice Loss & 61.73$\pm$0.94 & 92.40$\pm$0.40 & 31.00$\pm$1.31 \\
                         & $\lambda$Dice loss + (1\!-\!$\lambda$)CE loss ($\lambda$=0.7) & 62.53$\pm$0.64 & 93.00$\pm$0.20 & \textcolor{red}{32.06$\pm$1.50} \\
                         & ML (ours) & \textcolor{red}{62.60$\pm$0.34} & 93.33$\pm$0.30 & 31.66$\pm$0.41 \\
                         & CAML (ours) & 62.20$\pm$0.34 & 93.26$\pm$0.30 & 31.40$\pm$0.69 \\
\midrule
U\text{-}Net (1/4 Training data) & CE Loss & 59.20$\pm$0.34 & \textcolor{red}{93.33$\pm$0.11} & 25.00$\pm$0.34 \\
                         & Dice Loss & 60.93$\pm$1.36 & 92.06$\pm$0.75 & 29.66$\pm$1.92 \\
                         & $\lambda$Dice loss + (1\!-\!$\lambda$)CE loss ($\lambda$=0.7) & 60.20$\pm$0.34 & 92.53$\pm$0.46 & 28.06$\pm$0.90 \\
                         & ML (ours) & 60.33$\pm$0.30 & 93.13$\pm$0.11 & 27.80$\pm$0.60 \\
                         & CAML (ours) & \textcolor{red}{61.40$\pm$0.40} & 93.00$\pm$0.00 & \textcolor{red}{29.86$\pm$1.10} \\
\midrule
TransUNet (Full data) & CE Loss & 57.53$\pm$0.30 & \textcolor{red}{93.13$\pm$0.23} & 22.26$\pm$0.61 \\
                      & Dice Loss & 58.80$\pm$0.34 & 92.00$\pm$0.69 & \textcolor{red}{25.73$\pm$0.41} \\
                      & $\lambda$Dice loss + (1\!-\!$\lambda$)CE loss ($\lambda$=0.7) & 58.60$\pm$0.60 & 92.46$\pm$0.11 & 24.80$\pm$1.11 \\
                      & ML (ours) & 57.80$\pm$0.52 & 92.60$\pm$0.20 & 23.20$\pm$1.00 \\
                      & CAML (ours) & \textcolor{red}{59.13$\pm$0.70} & 93.00$\pm$0.20 & 25.33$\pm$1.11 \\
\midrule
TransUNet (1/2 Training data) & CE Loss & 56.80$\pm$0.60 & \textcolor{red}{93.20$\pm$0.52} & 20.33$\pm$1.20 \\
                             & Dice Loss & 58.20$\pm$0.69 & 91.53$\pm$0.80 & \textcolor{red}{25.13$\pm$0.80} \\
                             & $\lambda$Dice loss + (1\!-\!$\lambda$)CE loss ($\lambda$=0.7) & 57.00$\pm$1.44 & 91.80$\pm$0.91 & 22.33$\pm$2.08 \\
                             & ML (ours) & 57.20$\pm$0.40 & 92.73$\pm$0.11 & 21.60$\pm$0.72 \\
                             & CAML (ours) & \textcolor{red}{58.33$\pm$0.30} & 92.66$\pm$0.11 & 24.06$\pm$0.46 \\
\midrule
TransUNet (1/4 Training data) & CE Loss & 56.73$\pm$0.23 & \textcolor{red}{93.13$\pm$0.41} & 20.13$\pm$0.23 \\
                             & Dice Loss & \textcolor{red}{57.66$\pm$0.92} & 91.93$\pm$0.46 & \textcolor{red}{23.73$\pm$1.67} \\
                             & $\lambda$Dice loss + (1\!-\!$\lambda$)CE loss ($\lambda$=0.7) & 57.00$\pm$1.44 & 91.80$\pm$0.91 & 22.33$\pm$2.08 \\
                             & ML (ours) & 57.20$\pm$0.40 & 92.73$\pm$0.11 & 21.60$\pm$0.72 \\
                             & CAML (ours) & 56.66$\pm$0.50 & 91.93$\pm$0.80 & 21.40$\pm$1.11 \\
\midrule
ConvFormer+U\text{-}Net (Full data) & CE Loss & 60.80$\pm$0.00 & \textcolor{red}{93.73$\pm$0.11} & 27.73$\pm$0.11 \\
                             & Dice Loss & 62.26$\pm$0.50 & 93.03$\pm$0.47 & 32.80$\pm$1.40 \\
                             & $\lambda$Dice loss + (1\!-\!$\lambda$)CE loss ($\lambda$=0.7) & 61.73$\pm$1.36 & 93.00$\pm$0.52 & 30.53$\pm$2.33 \\
                             & ML (ours) & \textcolor{red}{63.93$\pm$0.64} & 93.66$\pm$0.23 & \textcolor{red}{34.46$\pm$1.10} \\
                             & CAML (ours) & 62.53$\pm$0.30 & 93.60$\pm$0.20 & 31.60$\pm$0.69 \\
\midrule
ConvFormer+U\text{-}Net (1/2 Training data) & CE Loss & 61.33$\pm$1.84 & \textcolor{red}{93.80$\pm$0.00} & 28.66$\pm$3.69 \\
                                     & Dice Loss & 61.46$\pm$1.17 & 92.66$\pm$0.11 & 30.46$\pm$2.20 \\
                                     & $\lambda$Dice loss + (1\!-\!$\lambda$)CE loss ($\lambda$=0.7) & 61.73$\pm$0.75 & 93.13$\pm$0.30 & 30.33$\pm$1.44 \\
                                     & ML (ours) & 62.00$\pm$0.52 & 93.60$\pm$0.20 & 30.53$\pm$0.90 \\
                                     & CAML (ours) & \textcolor{red}{63.80$\pm$1.20} & 93.66$\pm$0.11 & \textcolor{red}{33.73$\pm$2.10} \\
\midrule
ConvFormer+U\text{-}Net (1/4 Training data) & CE Loss & 60.60$\pm$1.03 & 93.40$\pm$0.52 & 27.66$\pm$2.53 \\
                                     & Dice Loss & 60.66$\pm$0.50 & 91.93$\pm$0.75 & 28.86$\pm$0.70 \\
                                     & $\lambda$Dice loss + (1\!-\!$\lambda$)CE loss ($\lambda$=0.7) & 60.86$\pm$1.20 & 92.73$\pm$0.41 & 29.20$\pm$2.22 \\
                                     & ML (ours) & 60.93$\pm$0.64 & 93.26$\pm$0.11 & 28.40$\pm$1.21 \\
                                     & CAML (ours) & \textcolor{red}{61.40$\pm$0.52} & \textcolor{red}{93.40$\pm$0.00} & \textcolor{red}{29.60$\pm$1.56} \\
\bottomrule
\end{tabular}
\end{table*}

\begin{table*}[t]
\centering
\tiny
\caption{Segmentation performance (IoU, \%) for different class and loss functions on the Drosophila dataset.}
\label{tab:Drosophilaseg}
\begin{tabular}{lccccccccc}
\toprule
Networks & Loss & mIoU & membrane  & mitochondria  & synapse  & glia/extracellular  & intracellular  \\
\midrule

U\text{-}Net (Full data) & CE Loss & 68.20$\pm$0.69 & 72.86$\pm$0.50 & 79.26$\pm$0.30 & 33.80$\pm$1.56 & 62.33$\pm$0.64 & 92.73$\pm$0.11 \\
                         & Dice Loss & 68.40$\pm$0.34 & 71.73$\pm$0.30 & 78.13$\pm$0.30 & 37.26$\pm$0.92 & 62.13$\pm$0.11 & 92.26$\pm$0.11 \\
                         & $\lambda$Dice loss + (1\!-\!$\lambda$)CE loss ($\lambda$=0.3) & 69.13$\pm$0.50 & \textcolor{red}{73.06$\pm$0.11} & 79.73$\pm$0.98 & 37.86$\pm$0.94 & \textcolor{red}{62.66$\pm$0.30} & \textcolor{red}{92.93$\pm$0.11} \\
                         & $\lambda$Dice loss + (1\!-\!$\lambda$)CE loss ($\lambda$=0.5) & 68.93$\pm$0.64 & 72.40$\pm$0.20 & 78.80$\pm$1.58 & 39.13$\pm$0.98 & 62.00$\pm$0.52 & 92.53$\pm$0.11 \\
                         & $\lambda$Dice loss + (1\!-\!$\lambda$)CE loss ($\lambda$=0.7) & 69.26$\pm$0.11 & 72.66$\pm$0.23 & 79.60$\pm$0.20 & \textcolor{red}{39.60$\pm$1.00} & 62.53$\pm$0.98 & 92.60$\pm$0.00 \\
                         & ML (ours) & \textcolor{red}{69.40$\pm$0.20} & 72.93$\pm$0.11 & \textcolor{red}{80.06$\pm$0.11} & 39.40$\pm$0.34 & 62.46$\pm$0.30 & 92.60$\pm$0.00 \\
                         & CAML (ours) & 69.20$\pm$0.52 & 72.53$\pm$0.11 & 79.40$\pm$1.11 & 39.26$\pm$0.94 & 62.60$\pm$0.20 & 92.73$\pm$0.30 \\
\midrule
U\text{-}Net (1/4 Training data) & CE Loss & 60.06$\pm$0.70 & \textcolor{red}{70.46$\pm$0.30} & 70.20$\pm$1.73 & 13.00$\pm$4.45 & 53.66$\pm$1.36 & 91.93$\pm$0.23 \\
                                 & Dice Loss & 62.86$\pm$0.41 & 69.73$\pm$0.41 & 69.40$\pm$0.60 & 28.93$\pm$0.90 & 54.66$\pm$1.00 & 91.46$\pm$0.11 \\
                                 & $\lambda$Dice loss + (1\!-\!$\lambda$)CE loss ($\lambda$=0.7) & 63.26$\pm$0.50 & 69.93$\pm$0.30 & 70.73$\pm$1.36 & 28.26$\pm$0.80 & 55.46$\pm$1.10 & 91.60$\pm$0.00 \\
                                 & ML (ours) & 63.66$\pm$0.41 & 70.40$\pm$0.34 & 72.13$\pm$0.70 & 28.66$\pm$0.94 & 55.60$\pm$1.31 & 91.93$\pm$0.11 \\
                                 & CAML (ours) & \textcolor{red}{64.06$\pm$0.23} & 70.20$\pm$0.20 & \textcolor{red}{73.00$\pm$2.30} & \textcolor{red}{29.13$\pm$0.90} & \textcolor{red}{56.33$\pm$0.11} & \textcolor{red}{92.00$\pm$0.20} \\
\midrule
U\text{-}Net (1/8 Training data) & CE Loss & 54.46$\pm$1.74 & 67.46$\pm$0.46 & 62.06$\pm$2.31 & 4.33$\pm$7.33 & 47.74$\pm$0.46 & 90.60$\pm$0.34 \\
                                & Dice Loss & 58.60$\pm$1.31 & 66.20$\pm$0.60 & 61.73$\pm$0.75 & 26.93$\pm$4.27 & 48.06$\pm$1.22 & 90.00$\pm$0.40 \\
                                & $\lambda$Dice loss + (1\!-\!$\lambda$)CE loss ($\lambda$=0.7) & 58.66$\pm$1.22 & 66.90$\pm$0.51 & 62.20$\pm$3.64 & 25.40$\pm$2.90 & 48.73$\pm$0.61 & 90.33$\pm$0.30 \\
                                & ML (ours) & 59.06$\pm$0.11 & 67.20$\pm$0.34 & 62.93$\pm$0.11 & 25.20$\pm$0.52 & 49.26$\pm$0.50 & 90.73$\pm$0.23 \\
                                & CAML (ours) & \textcolor{red}{60.13$\pm$0.41} & \textcolor{red}{69.70$\pm$3.65} & \textcolor{red}{64.20$\pm$1.38} & \textcolor{red}{27.40$\pm$2.62} & \textcolor{red}{50.26$\pm$0.57} & \textcolor{red}{90.80$\pm$0.20} \\
\midrule
TransUNet (Full data) & CE Loss & 65.26$\pm$0.98 & 71.60$\pm$0.20 & 72.46$\pm$1.33 & 32.86$\pm$4.27 & 56.86$\pm$1.33 & 92.20$\pm$0.00 \\
                      & Dice Loss & \textcolor{red}{66.93$\pm$0.11} & 71.53$\pm$0.23 & \textcolor{red}{74.86$\pm$0.98} & \textcolor{red}{38.46$\pm$1.02} & 58.53$\pm$0.94 & 92.00$\pm$0.20 \\
                      & $\lambda$Dice loss + (1\!-\!$\lambda$)CE loss ($\lambda$=0.7) & 67.13$\pm$0.30 & \textcolor{red}{71.80$\pm$0.34} & 74.66$\pm$0.92 & 38.20$\pm$1.31 & \textcolor{red}{58.66$\pm$0.70} & \textcolor{red}{92.26$\pm$0.11} \\
                      & ML (ours) & 66.40$\pm$0.40 & 71.60$\pm$0.00 & 72.80$\pm$0.40 & 38.06$\pm$1.33 & 57.13$\pm$0.70 & 92.20$\pm$0.00 \\
                      & CAML (ours) & 66.53$\pm$0.30 & 71.53$\pm$0.11 & 73.86$\pm$0.41 & 37.13$\pm$0.41 & 57.53$\pm$0.61 & \textcolor{red}{92.26$\pm$0.11} \\
\midrule
TransUNet (1/4 Training data) & CE Loss & 58.13$\pm$2.38 & 68.93$\pm$0.23 & 62.80$\pm$2.42 & 18.20$\pm$9.91 & 49.26$\pm$1.33 & 91.33$\pm$0.11 \\
                             & Dice Loss & 61.86$\pm$0.30 & 69.00$\pm$0.20 & \textcolor{red}{66.20$\pm$1.20} & 31.40$\pm$1.40 & 51.40$\pm$0.60 & 91.26$\pm$0.11 \\
                             & $\lambda$Dice loss + (1\!-\!$\lambda$)CE loss ($\lambda$=0.7) & \textcolor{red}{61.93$\pm$0.11} & \textcolor{red}{69.40$\pm$0.00} & 65.53$\pm$1.20 & 31.06$\pm$0.64 & \textcolor{red}{51.66$\pm$0.80} & 91.26$\pm$0.11 \\
                             & ML (ours) & 61.20$\pm$0.00 & 68.80$\pm$0.20 & 64.13$\pm$0.90 & \textcolor{red}{31.53$\pm$1.10} & 50.33$\pm$0.11 & \textcolor{red}{91.40$\pm$0.00} \\
                             & CAML (ours) & 60.93$\pm$0.23 & 68.66$\pm$0.11 & 63.60$\pm$0.00 & 31.33$\pm$0.11 & 49.60$\pm$0.69 & 91.20$\pm$0.00 \\
\midrule
TransUNet (1/8 Training data) & CE Loss & 49.33$\pm$0.80 & 64.86$\pm$0.11 & 48.13$\pm$2.60 & 3.60$\pm$3.27 & 40.53$\pm$1.40 & 90.06$\pm$0.30 \\
                             & Dice Loss & \textcolor{red}{57.46$\pm$0.46} & 65.80$\pm$0.60 & \textcolor{red}{55.66$\pm$0.75} & 29.26$\pm$0.94 & 46.46$\pm$0.75 & 90.26$\pm$0.30 \\
                             & $\lambda$Dice loss + (1\!-\!$\lambda$)CE loss ($\lambda$=0.7) & 57.40$\pm$0.69 & 66.06$\pm$0.30 & 54.40$\pm$1.24 & \textcolor{red}{30.20$\pm$0.20} & 46.06$\pm$0.98 & 90.26$\pm$0.30 \\
                             & ML (ours) & 57.40$\pm$0.20 & \textcolor{red}{66.40$\pm$0.40} & 53.93$\pm$0.98 & 30.00$\pm$0.34 & \textcolor{red}{46.06$\pm$0.23} & \textcolor{red}{90.46$\pm$0.30} \\
                             & CAML (ours) & 51.93$\pm$0.83 & 65.40$\pm$0.20 & 50.26$\pm$1.20 & 9.40$\pm$2.61 & 44.06$\pm$0.83 & 90.13$\pm$0.30 \\
\midrule

ConvFormer+U\text{-}Net (Full data) & CE Loss & 68.40$\pm$0.72 & 73.00$\pm$0.20 & 79.60$\pm$1.05 & 34.93$\pm$1.55 & 61.66$\pm$1.47 & \textcolor{red}{92.93$\pm$0.11} \\
                           & Dice Loss & 68.73$\pm$0.11 & 71.53$\pm$0.30 & 78.66$\pm$1.17 & 38.86$\pm$1.00 & \textcolor{red}{64.66$\pm$4.44} & 92.20$\pm$0.00 \\
                           & $\lambda$Dice loss + (1\!-\!$\lambda$)CE loss ($\lambda$=0.7) & 69.33$\pm$0.11 & 72.73$\pm$0.30 & 79.38$\pm$0.88 & 37.73$\pm$0.23 & 63.20$\pm$0.34 & 92.61$\pm$0.31 \\
                           & ML (ours) & 69.26$\pm$0.30 & \textcolor{red}{73.06$\pm$0.11} & \textcolor{red}{79.86$\pm$0.90} & 38.66$\pm$0.75 & 62.60$\pm$1.31 & \textcolor{red}{92.93$\pm$0.11} \\
                           & CAML (ours) & \textcolor{red}{69.53$\pm$0.46} & 72.93$\pm$0.23 & 79.46$\pm$0.61 & \textcolor{red}{39.20$\pm$1.00} & 63.40$\pm$0.80 & 92.80$\pm$0.00 \\
\midrule
ConvFormer+U\text{-}Net (1/4 Training data) & CE Loss & 60.86$\pm$0.23 & 70.40$\pm$0.34 & 72.53$\pm$1.36 & 15.38$\pm$3.12 & 53.46$\pm$1.00 & 92.00$\pm$0.20 \\
                                    & Dice Loss & 61.93$\pm$0.41 & 70.13$\pm$0.80 & 68.60$\pm$0.80 & 27.13$\pm$1.22 & 53.06$\pm$1.33 & 91.33$\pm$0.11 \\
                                    & $\lambda$Dice loss + (1\!-\!$\lambda$)CE loss ($\lambda$=0.7) & 63.06$\pm$0.75 & 70.20$\pm$0.20 & 70.40$\pm$2.00 & 27.93$\pm$1.22 & \textcolor{red}{55.33$\pm$0.83} & 91.80$\pm$0.20 \\
                                    & ML (ours) & 63.40$\pm$0.34 & 70.26$\pm$0.30 & 72.00$\pm$0.72 & 28.00$\pm$0.52 & 55.26$\pm$1.41 & 91.93$\pm$0.11 \\
                                    & CAML (ours) & \textcolor{red}{63.73$\pm$0.30} & \textcolor{red}{70.46$\pm$0.23} & \textcolor{red}{72.86$\pm$0.80} & \textcolor{red}{28.73$\pm$0.80} & 55.00$\pm$1.60 & \textcolor{red}{92.00$\pm$0.00} \\
\midrule
ConvFormer+U\text{-}Net (1/8 Training data) & CE Loss & 54.20$\pm$0.91 & 67.40$\pm$0.80 & 60.20$\pm$3.07 & 6.93$\pm$4.94 & 45.86$\pm$1.33 & 90.66$\pm$0.50 \\
                                    & Dice Loss & \textcolor{red}{59.60$\pm$0.20} & 66.93$\pm$0.23 & 61.33$\pm$2.31 & \textcolor{red}{27.93$\pm$0.83} & 47.20$\pm$1.58 & 90.26$\pm$0.11 \\
                                    & $\lambda$Dice loss + (1\!-\!$\lambda$)CE loss ($\lambda$=0.7) & 58.20$\pm$1.11 & 67.06$\pm$0.23 & 61.60$\pm$3.32 & 24.00$\pm$1.74 & 47.60$\pm$1.05 & 90.66$\pm$0.23 \\
                                    & ML (ours) & 58.66$\pm$0.23 & 66.93$\pm$0.30 & 63.33$\pm$0.30 & 23.20$\pm$1.24 & \textcolor{red}{49.06$\pm$1.11} & 90.46$\pm$0.11 \\
                                    & CAML (ours) & 58.80$\pm$0.00 & \textcolor{red}{67.40$\pm$0.20} & \textcolor{red}{63.40$\pm$0.72} & 25.33$\pm$2.34 & 47.80$\pm$0.60 & \textcolor{red}{90.73$\pm$0.11} \\
\bottomrule
\end{tabular}
\end{table*}

\subsection{Qualitative Results}
\label{43}

\begin{figure}[h]
    \centering
    \includegraphics[width=0.9\linewidth]{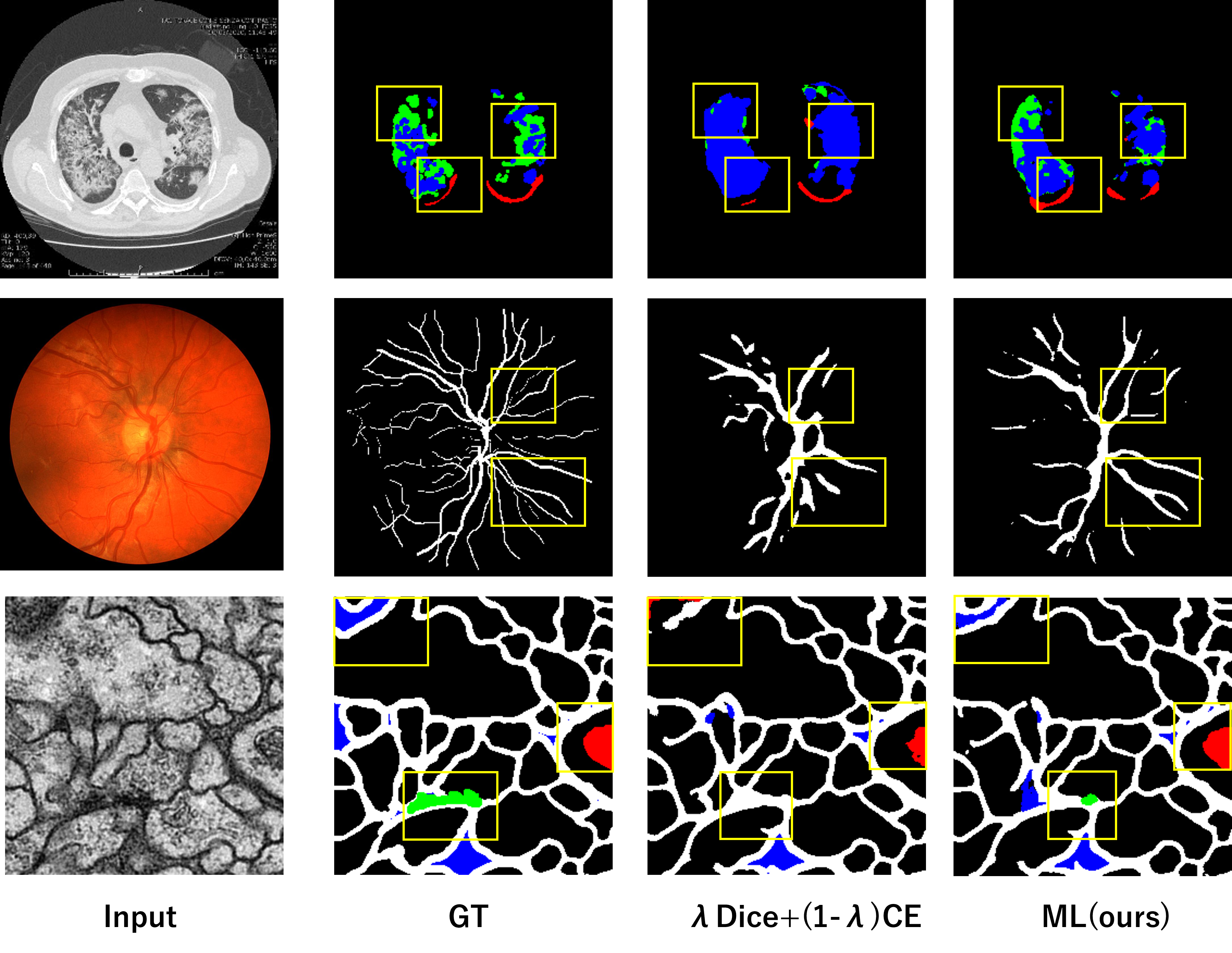}
    \caption{Qualitative comparison between our ML and a dataset-tuned additive loss function. Both methods used U-Net. The results are shown on three datasets: COVID-19 (row 1), CHASE (row 2), and Drosophila (row 3). Yellow boxes highlight regions where ML improved the missed or misclassified areas by additive loss.}
    \label{fig:MLkekka11}
\end{figure}

\begin{figure}[h]
    \centering
    \includegraphics[width=0.9\linewidth]{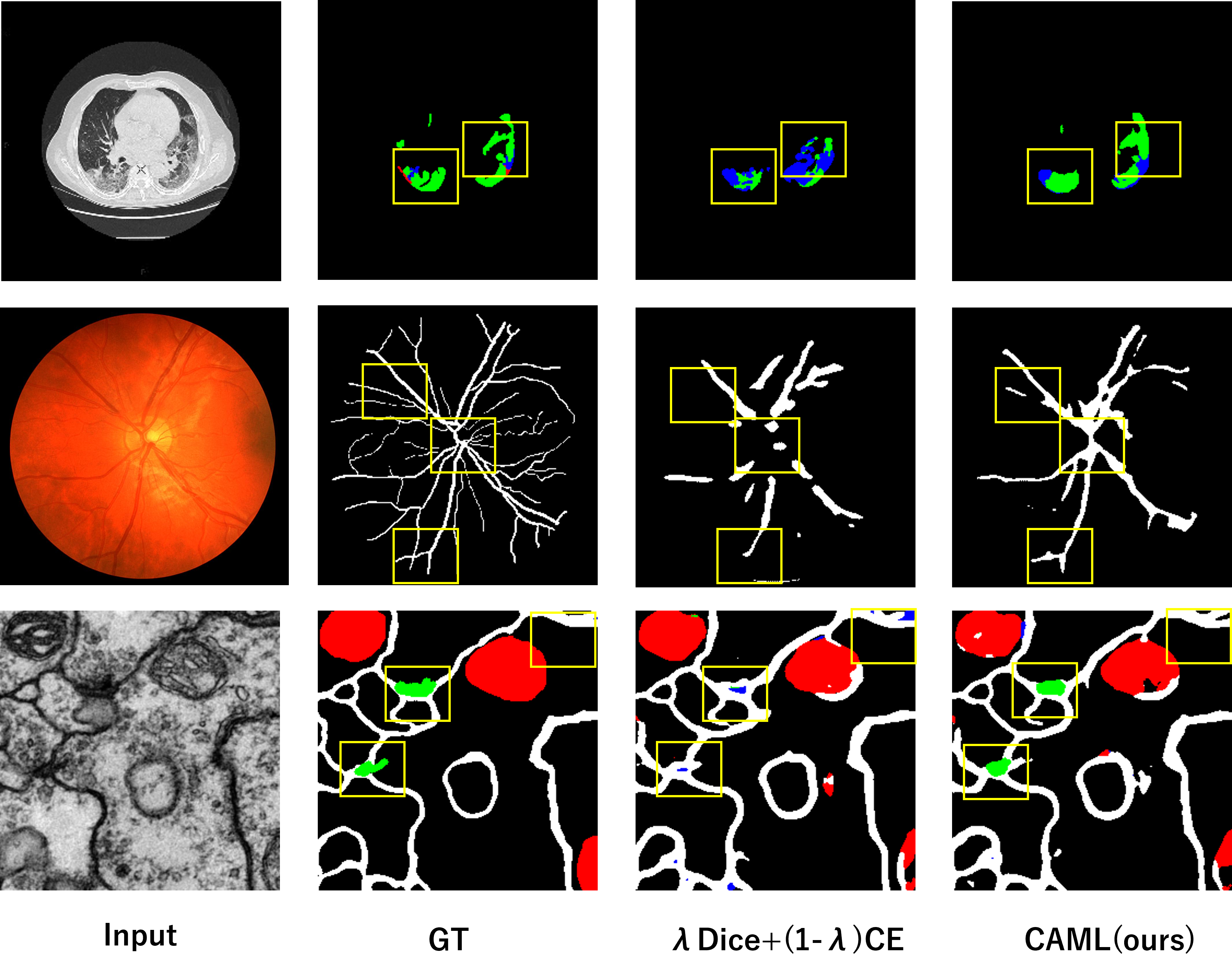}
    \caption{Qualitative comparison between the proposed CAML and a dataset-tuned additive loss function. Both methods used U-Net under a low-data regime (1/8 training samples). The results are shown on COVID-19 (row 1), CHASE (row 2), and Drosophila (row 3). Yellow boxes highlight regions where CAML outperforms the baseline in under- and over-segmented areas.}
    \label{fig:CAMLkekka18}
\end{figure}

\cref{fig:MLkekka11} presents qualitative comparisons between the proposed ML and a dataset-tuned additive loss function. Both method used U-Net.
Similarly, \cref{fig:CAMLkekka18} shows the results by the CAML, also based on the U-Net architecture, under a low-data regime in which only 1/8 of the original training samples are available. The yellow boxes in both Figures highlight the regions that were either missed or incorrectly segmented by the baseline method.
The first row of \cref{fig:MLkekka11} illustrates segmentation results on the COVID-19 dataset. Compared to the baseline, ML correctly identifies regions as ground glass opacity (green) that were previously misclassified as consolidations (blue, center-left). ML also improves the delineation of object boundaries (center-right) and enhances the detection of subtle pleural effusions (red, lower-left), which were under-segmented in the baseline.
The second row shows results on the CHASE dataset. ML successfully reconstructs disconnected retinal vessels (white, center-right) with improved topological continuity and suppresses false positives (bottom-right), resulting in a more complete vascular map.
The third row demonstrates results on the Drosophila dataset. ML significantly reduces false positives (upper-left) and achieves more accurate class-wise segmentation for mitochondria (red), membrane (white), and glia/extracellular structures (blue). Notably, ML increases the recall of mitochondria (center-right) and enables the detection of synapses (green, bottom-center) that were entirely missed by the baseline approach.

\cref{fig:CAMLkekka18} provides analogous comparisons under limited training data with the same U-Net.
On the COVID-19 dataset (first row), CAML improves segmentation of ground glass opacity (green, center-left) and corrects both under- and over-segmented areas (center-right), where baseline predictions were inconsistent or inaccurate.
In the CHASE dataset (second row), CAML recovers previously undetected retinal vessels (center) and reduces over-segmentation errors, particularly in challenging branched regions (center-left and bottom-center).
In the Drosophila dataset (third row), CAML enables the detection of synapses (green, center-left and left-center) that were previously missed, and corrects misclassified regions (upper-right) originally labeled as consolidations (blue), now correctly identified as intracellular structures.
These qualitative results demonstrate that ML consistently outperforms additive loss functions, even when the latter are optimized for each dataset. Furthermore, CAML, an extension of ML, retains high segmentation performance in extreme low-data settings, offering improved accuracy and robustness across diverse biomedical domains, while using the same U-Net architecture.




\subsection{Ablation study
}
\label{sec:ablation_alpha}

To evaluate the effectiveness of the dynamic scaling factor $\alpha = (1 - p)^D$ used in CAML, we conduct an ablation study by replacing the DICE score $D$ with a constant value $r \in \{1, 2, 3, 4, 5\}$. We use U-Net as the base architecture and perform experiments on three datasets: COVID-19, CHASE, and Drosophila under data-limited conditions: only 1/8 or 1/4 of the training images. The results are summarized in 
\cref{tab:ablation_miou}.

\begin{table}[htbp]
\centering
\scriptsize
\caption{Ablation study with constant $r$ values (instead of Dice score $D$) on mIoU.}
\begin{tabular}{c|c|c|c}
\hline
$r$ & COVID-19 & CHASE & Drosophila \\
\hline
1 & 37.70$\pm$1.02 & 61.20$\pm$1.03 & 59.40$\pm$1.11 \\
2 & 37.10$\pm$0.80 & 61.93$\pm$0.75 & 58.73$\pm$0.61 \\
3 & 37.60$\pm$0.40 & \textcolor{red}{62.60$\pm$0.40} & 58.93$\pm$0.64 \\
4 & 35.90$\pm$1.74 & 61.40$\pm$0.72 & 59.20$\pm$1.05 \\
5 & 37.80$\pm$0.61 & 61.06$\pm$0.64 & 59.07$\pm$0.80 \\
\hline
\textbf{CAML (Ours)} & \textcolor{red}{38.06$\pm$0.41} & 61.40$\pm$0.40 & \textcolor{red}{60.13$\pm$0.41} \\
\hline
\end{tabular}
\label{tab:ablation_miou}
\end{table}

These results demonstrate that employing a constant $r$ in the scaling factor $alpha$ leads to dataset-dependent and inconsistent performance. For instance, while the CHASE dataset achieves its best performance at $r=3$, the same parameter setting results in degraded performance on the COVID-19 and Drosophila datasets.
In contrast, the full formulation of CAML, which dynamically adjusts $alpha$ based on both prediction confidence $p$ and the overall DICE score $D$, consistently attains the highest or comparable mIoU across all datasets.
This adaptive scaling mechanism enables more effective emphasis on difficult samples, particularly under limited data conditions, and yields lower standard deviations in mIoU compared to fixed $r$ settings, indicating improved training stability.
Overall, these findings highlight that the DICE-based adaptive scaling strategy of CAML obviates the need for dataset-specific tuning, offering a more robust and generalizable approach for diverse segmentation tasks.




\subsection{Effectiveness of ML and CAML}
\label{sec5}

We propose 
ML and CAML as alternatives to conventional additive loss functions. ML multiplies CE loss and Dice loss, enabling:
\begin{itemize}
    \item Amplified gradients for misclassified or low-confidence samples, accelerating learning.
    \item Suppressed gradients for high-confidence samples, reducing overfitting.
\end{itemize}

This simple yet effective formulation avoids hyperparameter tuning and achieves stable performance across datasets. The notable mIoU gain on CHASE using ConvFormer+U-Net supports this, especially in binary segmentation where prediction biases are amplified, enhancing gradient contrast compared to additive losses (\cref{multihikaku}).
CAML further introduces exponential weighting based on pixel-wise prediction confidence and the global Dice coefficient. This dynamically emphasizes uncertain predictions and down-weights confident ones:
\begin{itemize}
    \item Early-stage or difficult samples receive stronger gradients.
    \item Confident predictions are gradually suppressed, preventing overfitting.
\end{itemize}

CAML particularly improves generalization in low-data settings and surpasses ML when prediction distributions are less extreme, as gradients behave similarly to ML while maintaining regularization (\cref{camlhikaku}).
However, on TransUNet, mIoU degrades with CAML. Due to its reliance on global attention, TransUNet underperforms on fine structures and boundaries, lowering prediction confidence\cite{transUnet},\cite{TransUnetakka1}. This results in excessive loss amplification and sensitivity to noise under sample-scarce conditions. Furthermore, its attention-based representation lacks the robustness in capturing local details compared to CNNs\cite{transUnet},\cite{transUnetakka2}, increasing susceptibility to CAML-induced overfitting.
Therefore, applying CAML to Transformer-based models may require auxiliary regularization or smoothing to stabilize confidence distributions. Loss scaling mechanisms should be tailored to model architecture.
In summary, ML and CAML are effective loss designs that dynamically adjust learning signals, improving segmentation performance and mitigating overfitting, especially in medical imaging tasks with limited data.

\section{Conclusion}
\label{concul}

We propose Multiplicative Loss (ML) and its extension, Confidence-Adaptive Multiplicative Loss (CAML), which combine Dice Loss and Cross Entropy Loss to amplify gradients for misclassified samples while suppressing those for high-confidence predictions. ML enhances the performance in CNN-based segmentation models without hyperparameter tuning. CAML further improves stability and accuracy by dynamically adjusting the scaling factor based on prediction confidence and the overall Dice score.

Experiments on multiple datasets show that both ML and CAML consistently outperform conventional additive losses in mean IoU. Ablation studies confirm the effectiveness of CAML's confidence-based scaling. However, their performance drops on Transformer-based models like TransUNet, likely due to less discriminative features from self-attention’s uniform global aggregation\cite{transUnet},\cite{TransUnetakka1}, indicating structural incompatibility with our loss design. Future work will refine these losses and extend them to large-scale models and diverse segmentation tasks.






\section*{Erratum}

There was an mistake in \cref{3.2}. 
As a result, the entire \cref{3.2} has been updated to reflect this correction. 
In particular, in the Ablation Study, all occurrences of the pixel-level prediction probability $p$ have been replaced with the batch-level average prediction probability $\bar{p}$. 
This correction does not affect the main conclusions or the experimental results of the paper.

\begin{figure*}[t]
    \centering
    \includegraphics[width=\textwidth]{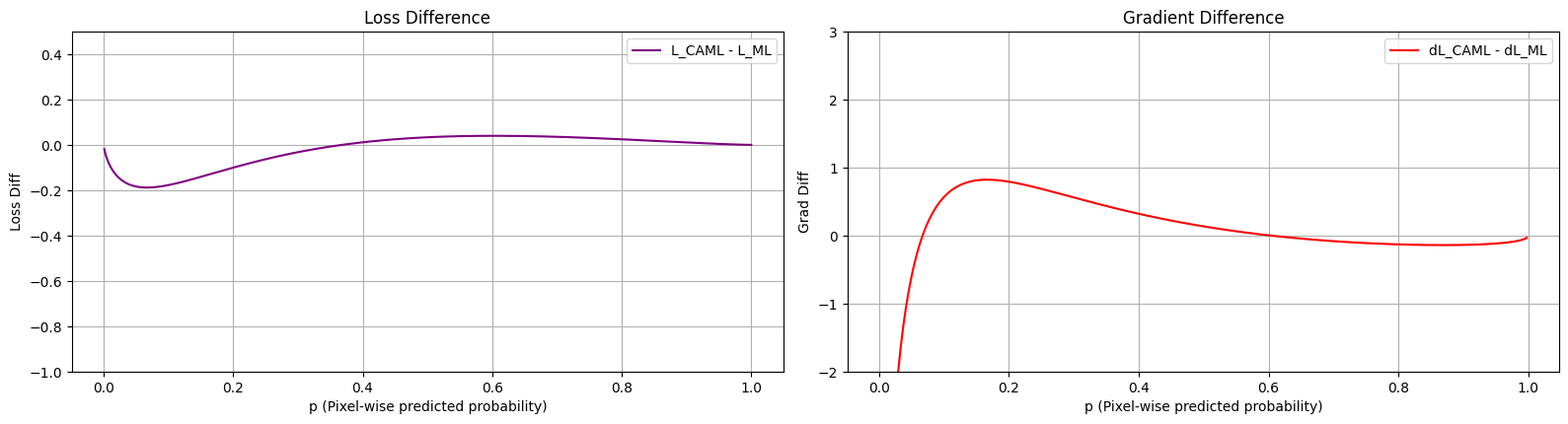}
    \caption{Comparison of gradient magnitudes between CAML and ML with respect to the predicted probability $p$. 
    CAML adaptively adjusts gradient magnitudes based on instance-level confidence, giving higher gradient updates to hard samples while suppressing excessive updates for high-confidence predictions.}
    \label{camlhikakuv2}
\end{figure*}

For clarity, we reproduce the corrected version of the section below. 
In addition, the corresponding Figure \cref{camlhikaku} has been updated and is now shown as \cref{camlhikakuv2}.
\\

In this section, we extend the proposed ML introduced in \cref{eq:ml_loss} by incorporating an adaptive weighting mechanism that dynamically adjusts the loss contribution based on the average prediction probability across all pixels and classes in the batch through training. This extension, called CAML, is defined as
{\footnotesize
\begin{equation}
    \mathcal{L}_{\mathrm{CAML}} = \mathcal{L}_{\mathrm{Dice}} \cdot \left( \mathcal{L}_{\mathrm{CE}} \right)^{\alpha}, \quad \text{where} \quad \alpha = (1 - \bar{p})^D,
    \label{eq:CAML_compact_batch}
\end{equation}
}
where $\mathcal{L}_{\mathrm{Dice}}$ and $\mathcal{L}_{\mathrm{CE}}$ denote the Dice loss and Cross Entropy loss (see \cref{eq:dice_loss_2} and \cref{eq:ce_loss}), respectively. $\bar{p}$ represents the average predicted probability across all pixels and classes in the batch, and $D$ is the Dice coefficient in \cref{eq:dice_loss_1} indicating segmentation consistency.

Note that $\bar{p}$ is a single scalar for the batch and does not vary across pixels. However, by multiplying with the Dice coefficient, which depends on pixel-wise predictions, the effective gradient is modulated at the pixel level, allowing the model to focus on more uncertain or difficult regions.

The gradient of $\mathcal{L}_{\mathrm{CAML}}$ with respect to $\bar{p}$ is derived via the chain rule as
{\footnotesize
\begin{equation}
\begin{aligned}
    \frac{\partial \mathcal{L}_{\mathrm{CAML}}}{\partial p_{i,c}} 
    =\,& \left(\mathcal{L}_{\mathrm{CE}}\right)^{\alpha} \Biggl[ 
    \frac{\partial \mathcal{L}_{\mathrm{Dice}}}{\partial p_{i,c}} \\
    & + \mathcal{L}_{\mathrm{Dice}} \cdot \left(
    \frac{\partial \alpha}{\partial p_{i,c}} \log \mathcal{L}_{\mathrm{CE}} 
    + \alpha \cdot \frac{1}{\mathcal{L}_{\mathrm{CE}}} \frac{\partial \mathcal{L}_{\mathrm{CE}}}{\partial p_{i,c}}
    \right) 
    \Biggr],
\end{aligned}
\label{eq:CAML_grad_compact_batch}
\end{equation}
}
where
{\footnotesize
\begin{equation}
\frac{\partial \alpha}{\partial p_{i,c}} 
= \frac{\partial \alpha}{\partial \bar{p}} \cdot \frac{\partial \bar{p}}{\partial p_{i,c}}
= \frac{\partial \alpha}{\partial \bar{p}} \cdot \frac{1}{N}
= \frac{-D (1 - \bar{p})^{D-1}}{N}.
\label{eq:alpha_deriv_pixel_chain}
\end{equation}
}

By incorporating the Dice coefficient into the exponent $\alpha$, CAML dynamically modulates supervision based on the current batch average prediction probability. As training progresses and both $\bar{p}$ and $D$ increase, the weight on $\mathcal{L}_{\mathrm{CE}}$ decreases exponentially, effectively reducing overfitting on easy samples while emphasizing harder cases.

Moreover, CAML adaptively scales gradient magnitudes based on the batch-average prediction probability $\bar{p}$ and overall batch performance. In CAML, $\bar{p}$ is modulated at the pixel level by the Dice coefficient, allowing the model to focus on uncertain or difficult regions while minimizing unnecessary updates and facilitating stable convergence.


Importantly, this adaptive mechanism is particularly effective when training data is limited. In such cases, conventional additive or simple multiplicative losses tend to overfit easy samples due to the scarcity of training data. CAML mitigates this by prioritizing harder samples, thereby balancing the bias-variance trade-off and achieving better generalization even with small datasets.

CAML’s effect stems from the adaptive weighting mechanism governed by the batch-level average prediction probability $\bar{p}$ and the global Dice coefficient $D$, which together determine the exponent $\alpha = (1 - \bar{p})^D$. As training progresses and segmentation performance improves, $\alpha$ decreases for batches with high average $\bar{p}$, reducing their influence on the overall loss. This dynamic attenuation via the Dice coefficient effectively suppresses gradient updates from already well-learned easy pixels, allowing the model to focus on more uncertain or challenging regions. Such targeted learning prevents overfitting to redundant information and noise, which are common issues in limited data regimes, and instead promotes better feature abstraction by focusing model capacity on informative errors. Additionally, the small but non-zero gradients in high-average-confidence batches serve as a form of implicit regularization, contributing to optimization stability and further enhancing generalization.

In summary, CAML enables dynamic loss adaptation aligned with training dynamics at the batch level, promoting robust optimization even under limited data scenarios. Experiments on multiple medical image segmentation benchmarks demonstrated that CAML consistently outperforms conventional additive and multiplicative loss formulations, achieving superior accuracy, enhanced stability, and improved the robustness to overfitting. These results establish CAML as an effective loss function for small-scale medical image learning tasks.


{
    \small
    \bibliographystyle{ieeenat_fullname}
    \bibliography{main}
}

\end{document}